%% file: toposheaf_iclr2027.tex
\documentclass{article}

\usepackage{iclr2027_conference,times}
\input{math_commands.tex}

\usepackage[utf8]{inputenc}
\usepackage[T1]{fontenc}
\usepackage{microtype}
\usepackage{graphicx}
\usepackage{booktabs}
\usepackage{amsmath}
\usepackage{amssymb}
\usepackage{amsfonts}
\usepackage{amsthm}
\usepackage{algorithm}
\usepackage{algorithmic}
\usepackage{multirow}
\usepackage{subcaption}
\usepackage{wrapfig}
\usepackage{xcolor}
\usepackage{colortbl}
\usepackage{bm}
\usepackage{xspace}
\usepackage{hyperref}
\usepackage{url}

\newcommand{\method}{\textsc{ToCoMAS}\xspace}

\title{Topological Coherence for Self-Evolving Multi-Agent Systems}

\renewcommand{\thefootnote}{\fnsymbol{footnote}}
\author{
\textbf{Sen Zhao\textsuperscript{1}\thanks{Equal contribution.} \quad
Ruiqi Kong\textsuperscript{1}\footnotemark[1] \quad
Zuyu Zhang\textsuperscript{1} \quad
Lifeng Shen\textsuperscript{2}} \\
\textbf{Xinyu He\textsuperscript{4} \quad
Xu Zhang\textsuperscript{1}\thanks{Corresponding author: \texttt{zhangx@cqupt.edu.cn}.} \quad
Qinghua Zhang\textsuperscript{3}} \\[0.35em]
{\normalfont\small \textsuperscript{1}Academy of Advanced Interdisciplinary Studies,} \\
{\normalfont\small Chongqing University of Posts and Telecommunications, Chongqing, China} \\
{\normalfont\small \textsuperscript{2}School of Artificial Intelligence,} \\
{\normalfont\small Chongqing University of Posts and Telecommunications, Chongqing, China} \\
{\normalfont\small \textsuperscript{3}School of Computer Science and Technology,} \\
{\normalfont\small Chongqing University of Posts and Telecommunications, Chongqing, China} \\
{\normalfont\small \textsuperscript{4}Towngas, China}
}

\iclrfinalcopy

\begin{document}
\maketitle
\renewcommand{\thefootnote}{\arabic{footnote}}
\setcounter{footnote}{0}
\lhead{Under review}

\begin{abstract}
  Complex tasks inherently couple workflow structure, agent responsibility,
  collaboration, and memory access: task regions delimit responsibility and
  tool scope, cross-region dependencies give rise to handoffs, and ownership
  boundaries delimit private and selectively shared memory.
  {Existing methods can jointly optimize agent and communication
  structures, yet such optimization does not by itself require responsibility,
  handoff, and memory boundaries to remain consistent with task dependencies.
  We term this requirement \emph{topological coherence}.}
  {We introduce \method, a \emph{Topology-Coherent Multi-Agent
  System}.} \method grounds a task graph in tool interfaces, organizes compatible
  task nodes into reusable responsibility domains, and derives
  dependency-induced and profile-conditioned collaboration together with
  boundary-regulated memory visibility. {During online self-evolution, \method proposes coupled changes to agent, collaboration, and memory policies, retaining for subsequent tasks only candidates that satisfy structural constraints and improve evaluated reward.} Across BBEH,
  WorkBench, SWE-Bench-Verified, and CoMemBench, \method improves task success
  over baselines across backbones. CoMemBench also shows {gains
  over the self-evolving baseline} in verified
  progress, handoffs, and memory isolation.
\end{abstract}

\input{Sections/Introduction}
\input{Sections/RelatedWork}
\input{Sections/Preliminaries}
\input{Sections/Method}
\input{Sections/Experiments}
\input{Sections/Conclusion}

\subsection*{Reproducibility Statement}
\input{Sections/Reproducibility}

\subsection*{AI Use Statement}
\input{Sections/AIUse}

\bibliographystyle{iclr2027_conference}
\bibliography{reference}

\clearpage
\appendix
\input{Sections/Appendix}

\end{document}

%% file: math_commands.tex
\usepackage{amsmath,amsfonts,bm}

\def\eqref#1{equation~\ref{#1}}

\def\1{\bm{1}}

\DeclareMathAlphabet{\mathsfit}{\encodingdefault}{\sfdefault}{m}{sl}
\SetMathAlphabet{\mathsfit}{bold}{\encodingdefault}{\sfdefault}{bx}{n}



%% file: Sections/Introduction.tex
\section{Introduction}

Large language model (LLM) based multi-agent systems (MAS) provide a general
framework for combining specialized reasoning, tool use, and interaction
policies~\citep{guo2024large,li2024survey}. Conversational and role-based systems
such as CAMEL, AutoGen, and AgentVerse~\citep{li2023camel,wu2024autogen,
chen2024agentverse}, together with peer review, debate, MetaGPT, and ChatDev
~\citep{xu2023towards,du2023improving,hong2023metagpt,qian2024chatdev}, make
increasingly complex tasks accessible. This progress also turns organization
into a central design problem: a MAS must decide which task regions each agent
owns, which tools and information it may access, where verification is required,
and how local outputs compose into a globally executable process.

Recent work automates these choices by searching computation graphs, agentic
programs, or workflows~\citep{zhuge2024gptswarm,hu2024automated,
zhang2024aflow}; adapting communication topology or task orchestration
~\citep{liu2024dynamic,zhang2025gdesigner,yang2025agentnet,wu2026card,
ke2026masorchestra}; and retaining structured or decentralized experience
~\citep{zhang2025gmemory,wang2026emem,yang2026plugmem,hao2026decentmem}.
Evolutionary methods further generate agents~\citep{yuan2025evoagent}. The most
comprehensive system-level formulation jointly evolves agent roles, prompts,
models, tools, and communication topology, while retaining a candidate pool and
experience memory for cross-task reuse~\citep{hu2026evomas}. Nevertheless, task
workflows, agent responsibilities, collaboration, and memory remain separately
represented structures, without an explicit constraint that keeps their
relations compatible as the system changes.

\begin{wrapfigure}{r}{0.50\textwidth}
  \vspace{-0.5\baselineskip}
  \centering
  \includegraphics[width=\linewidth]{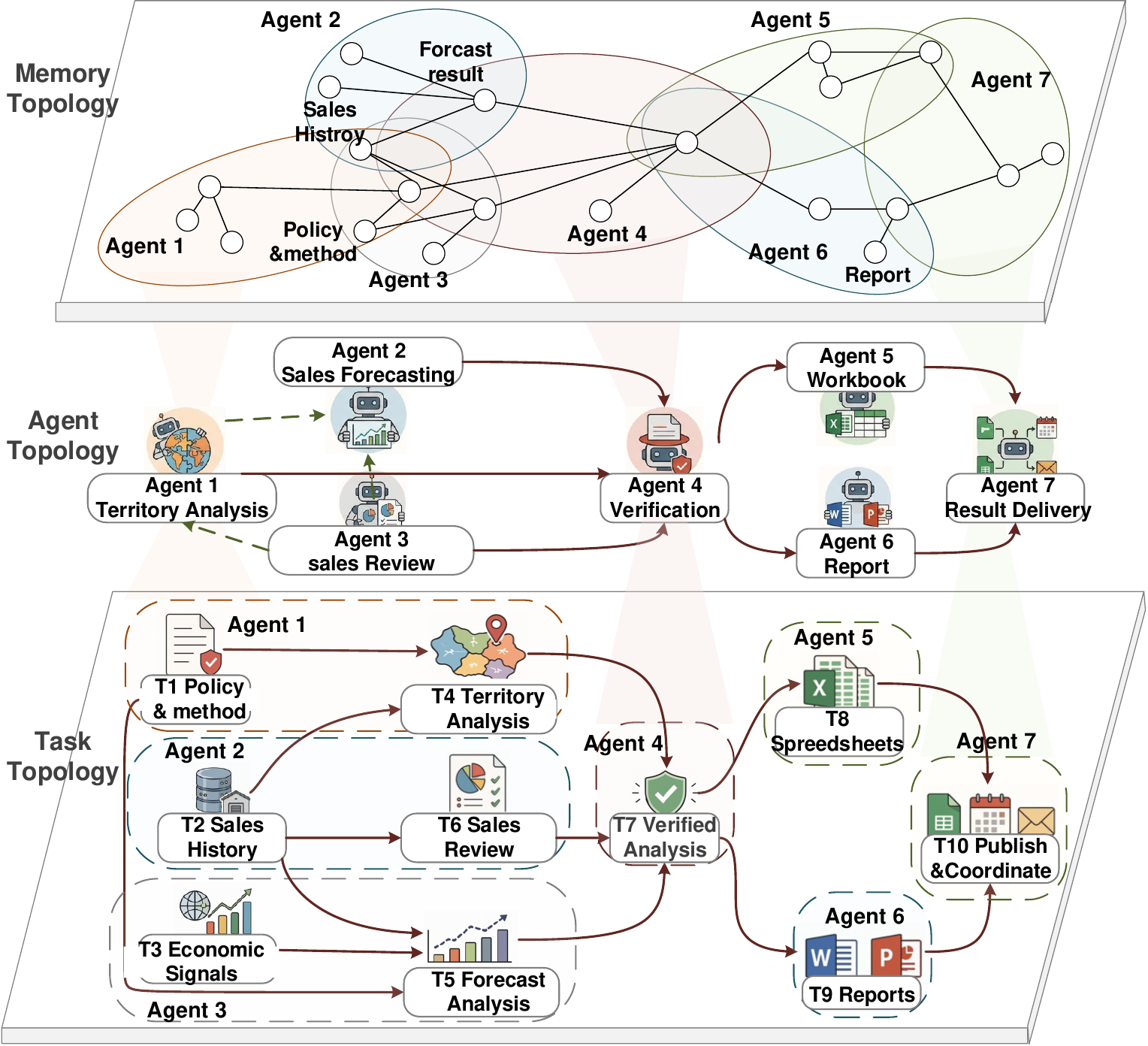}
  \caption{\textbf{Topological coherence in ToCoMAS.} Task topology jointly
  induces agent responsibilities, collaboration interfaces, and memory
  boundaries, which evolve as a coupled organization.}
  \label{fig:motivation}
  \vspace{-0.5\baselineskip}
\end{wrapfigure}

Figure~\ref{fig:motivation} illustrates this relation on a query whose policy,
sales-history, and economic-evidence branches feed workbook, reporting, and
delivery tasks. The workflow itself supplies a common organizational reference:
task regions delimit responsibility and tool scope, cross-region dependencies
create handoffs and optional verification, and ownership boundaries delimit
private and selectively shared memory. These are different views of one
task-conditioned organization. We call their agreement \emph{topological
coherence}: not similarity between graph shapes, but consistency among the
relations each layer derives from the task topology~\citep{bodnar2021weisfeiler}.

Building on this formulation, we propose \textbf{\method}, a
\emph{Topology-Coherent Multi-Agent System} that carries this constraint through
organization and self-evolution. \method maps an upstream-provided task graph to
a public tool graph under global operation and dependency constraints, then groups
structurally and operationally compatible regions into reusable agent responsibility domains, allowing one agent to own
multiple related subgraphs while separating heterogeneous execution semantics.
It then derives collaboration and memory coordination from relations among these
domains: responsibility ownership delimits local tools and private memory,
task-induced adjacency provides dependency-aware handoffs, and boundary policies
regulate selective sharing and verifier escalation. {Finally, an
online controller proposes changes to reusable structural policies and
compiles each candidate against the task topology. It reconciles affected
ownership, handoff, and memory relations, then retains structurally valid
candidates only when their evaluated reward improves on the parent.} We
evaluate fixed and evolving
variants against a direct-call control, prespecified and automatically designed
baselines, and a self-evolving baseline across reasoning, tool-use, and
software-engineering benchmarks. Our contributions are:
\begingroup
\setlength{\leftmargini}{2em}
\begin{itemize}
  \item We formulate \emph{topological coherence} as a structural constraint
  that keeps task workflows, agent responsibilities, collaboration interfaces,
  and memory boundaries consistent during self-evolution.
  \item We develop a topology-grounded architecture in which task and tool
  relations induce reusable agent responsibility domains, dependency-aware and
  profile-conditioned collaboration, and boundary-regulated memory access;
  execution feedback guides coupled structural proposals.
  \item We establish controlled comparisons with prespecified, automatically
  designed, and self-evolving MAS baselines to separate architectural gains
  from evolutionary search.
\end{itemize}
\endgroup

%% file: Sections/RelatedWork.tex
\section{Related Work}

Early role-based and conversational MAS organize collaboration through role play,
proposal aggregation, debate, and peer review
~\citep{li2023camel,wu2024autogen,chen2024agentverse,du2023improving,
xu2023towards}; MetaGPT and ChatDev encode specialized roles in manually
specified software-development workflows~\citep{hong2023metagpt,qian2024chatdev}.
Automatic design searches optimizable computation graphs (GPTSwarm)
~\citep{zhuge2024gptswarm}, code-defined agentic systems and workflow programs
(ADAS and AFlow)~\citep{hu2024automated,zhang2024aflow}, or agents, prompts,
and communication structures (AutoAgents, MASS, and ARG-Designer)
~\citep{chen2023autoagents,zhou2026mass,li2026argdesigner}. MAS-Orchestra
adds holistic task-conditioned orchestration~\citep{ke2026masorchestra}.
These methods move beyond fixed protocols but favor atomic roles or workflow
nodes over reusable task-and-tool responsibility domains.

Dynamic agent networks adapt participating agents at inference time
~\citep{liu2024dynamic}; G-Designer learns task-aware communication graphs,
AgentNet evolves decentralized coordination, and CARD conditions communication
topology on dynamic environments
~\citep{zhang2025gdesigner,yang2025agentnet,wu2026card}. Graph-conditioned
language models and graph retrieval guide reasoning and information selection
~\citep{tang2024graphgpt,he2024gretriever}, while topological learning studies
local-to-global compatibility through shared incidence and boundaries
~\citep{hansen2019spectral,bodnar2021weisfeiler}. In memory organization,
G-Memory stores hierarchical collaboration experience, E-mem reconstructs
episodic context through local assistants, and PlugMem organizes reusable
knowledge in a typed graph
~\citep{zhang2025gmemory,wang2026emem,yang2026plugmem}. Together, these lines
support relational reasoning, communication, and memory; they do not, by
themselves, jointly derive task ownership, tool authority, required handoffs,
and memory visibility from a shared task-and-tool topology.

Recent self-evolution methods revise organizations using feedback: EvoAgent
expands agent populations through mutation and crossover, while DecentMem uses
decentralized private memories for agent-level self-evolution
~\citep{yuan2025evoagent,hao2026decentmem}. EvoMAS, the closest system-level
approach, represents a MAS as a structured configuration and jointly evolves
agent roles, prompts, model and tool assignments, and communication topology;
it also retains a candidate pool and experience memory for cross-task reuse
~\citep{hu2026evomas}. Yet co-editing these components does not itself require
changes in responsibility to induce compatible agent capabilities, handoffs,
and memory permissions after a structural change. \method instead enforces
\emph{topological coherence}, deriving agent responsibilities, required
handoffs, and memory boundaries from shared task-and-tool regions.

%% file: Sections/Preliminaries.tex
\section{Problem Formulation}

Following a configuration-based formulation for evolutionary multi-agent
systems~\citep{hu2026evomas}, a multi-agent system is specified by:
\begin{equation}
C=\left(G,\mathcal{A},\mathcal{V}_{\mathrm{in}},
\mathcal{V}_{\mathrm{out}}\right),
\label{eq:mas-configuration}
\end{equation}
where $G=(\mathcal{V},\mathcal{E})$ is a directed communication graph over
agents, $\mathcal{A}=\{A_i\}_{i=1}^{k}$ is the corresponding set of agent
configurations, and $\mathcal{V}_{\mathrm{in}},
\mathcal{V}_{\mathrm{out}}\subseteq\mathcal{V}$ denote the input and output
agents. Each $A_i=(b_i,p_i,\Gamma_i)$ contains a backbone model, prompt, and
accessible tool set. {Given a task $q\in\mathcal{Q}$ with task
graph $T_q=(\mathcal{U}_q,\mathcal{D}_q)$, let
$H_q=(\mathcal{U}_q,\mathcal{D}_q,\Psi_q)$ be its tool-grounded executable
topology.} The responsibility map
$\pi_q:\mathcal{U}_q\rightarrow\mathcal{S}_q$ groups task nodes into regions,
and the ownership map $\phi_q:\mathcal{S}_q\rightarrow\mathcal{V}$ assigns
these regions to capable agents. The task-conditioned configuration is:
\begin{equation}
{C_q=\left(C,H_q,\pi_q,\phi_q,\Lambda_q,\mathcal{M}_q\right),
\qquad
\overline{H}_q=H_q/\pi_q.}
\label{eq:topological-coherence}
\end{equation}
{Here $C$ is instantiated with the task-conditioned collaboration
graph $G_q$. The policy $\Lambda_q$ governs memory admission and visibility,
whereas $\mathcal{M}_q$ denotes runtime memory records. We call $C_q$
topology-coherent when responsibility regions have capable owners, required
cross-region dependencies have compatible handoffs, and memory visibility
respects the same ownership boundaries.} Executing $C_q$ on $q$ yields an output
$y_q$ and trace $\tau_q$; configuration quality is evaluated by:
\begin{equation}
\begin{aligned}
(y_q,\tau_q)&=\operatorname{Exec}(C_q,q),\\
R(q,C_q)&=\operatorname{Metrics}(q,C_q)
-\beta\cdot\operatorname{Cost}(C_q).
\end{aligned}
\label{eq:configuration-reward}
\end{equation}
{Here} $\operatorname{Metrics}(q,C_q)$ measures accepted execution quality and
may incorporate bounded evidence of useful memory context, while
{$\operatorname{Cost}(C_q)$ denotes the measured resource cost
used in candidate selection.} The scalar $\beta\geq0$
controls the performance--efficiency trade-off. The trace provides feedback for
structural proposals; candidates are retained only when they pass executable
structural checks and improve accepted reward. Accumulated traces inform
subsequent evolution and selection.

%% file: Sections/Method.tex
\section{The Proposed Model}
\label{sec:method}

\begin{figure*}[t]
  \centering
  \includegraphics[width=\textwidth]{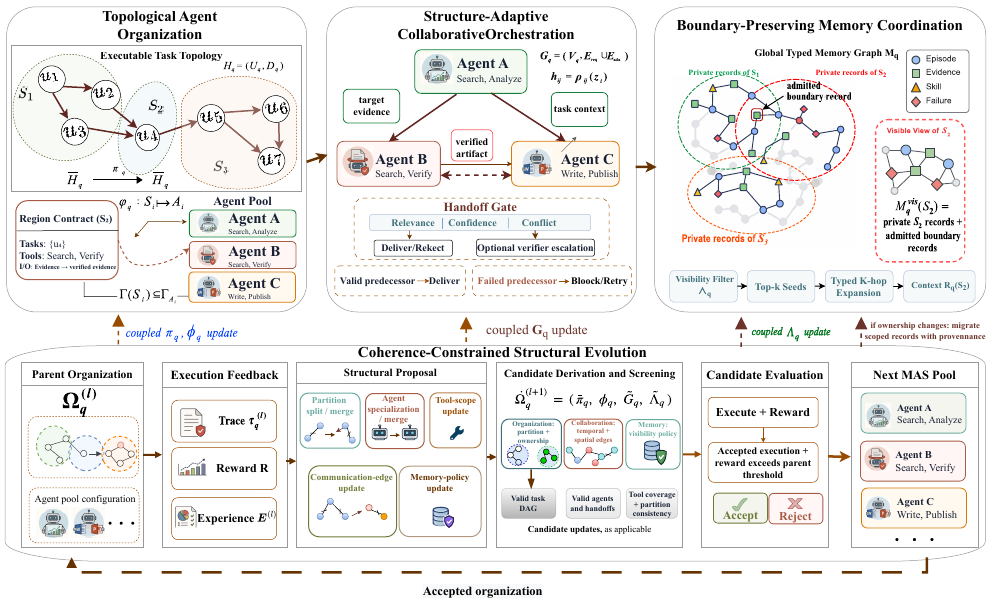}
  \caption{\textbf{Overview of \method.} Task-derived responsibility regions
  organize agents, handoffs, and memory boundaries; online evolution preserves
  their coherence.}
  \label{fig:tocomass-framework}
\end{figure*}

We present \textbf{\method}, a topology-coherent framework for constructing
and evolving multi-agent systems. Its central principle is that agent
organization, collaboration, and memory should not be designed as independent
configuration choices. Instead, they are derived from a common task topology
and jointly screened as the system evolves. Given a task graph $T_q$
and the public tool graph $K$, \method first grounds every existing task node by
a whole-graph map $\Psi_q$, without redecomposing the user query. It then forms a
quotient topology over responsibility regions, assigns these regions to capable
agents, and derives communication and memory boundaries from the same maps. This
construction is summarized by:
\begin{equation}
{T_q\xrightarrow{\ \Psi_q\ }H_q
\xrightarrow{\ \pi_q\ }
\overline{H}_q,
\qquad
\phi_q:\mathcal{S}_q\longrightarrow\mathcal{V}_q.}
\label{eq:method-construction}
\end{equation}
{Here $\pi_q$ groups task nodes into responsibility regions,
$\overline{H}_q$ preserves their cross-region dependencies, and $\phi_q$
assigns regions to agents. These maps determine required handoffs and constrain
the collaboration graph $G_q$ and memory policy $\Lambda_q$; additional
communication and retrieved records remain subject to their respective
policies.} As illustrated in Figure~\ref{fig:tocomass-framework}, the four
modules below realize this construction and preserve it during structural
evolution.

\subsection{Topological Agent Organization}
\label{sec:agent-organization}

\paragraph{Task topology and responsibility quotient.}
For a task $q$, let $T_q=(\mathcal{U}_q,\mathcal{D}_q)$ be the given task graph
and let $K=(\mathcal{T},\mathcal{R})$ be a public graph of tool interfaces and
their operated resources. Each node exposes an operation requirement
$R(u)$---including operations, resources, effects, receipt types, and declared
input--output semantics. The grounding map
$\Psi_q:T_q\rightarrow 2^{\mathcal{T}}$ selects a minimal valid tool bundle
$\Gamma(u)=\Psi_q(u)$ for every existing node jointly, subject to capability
coverage, edge-semantic compatibility, side-effect constraints, and provider
continuity. Gold trajectories and evaluator state are excluded from $\Psi_q$.
The grounded executable topology is $H_q=(\mathcal{U}_q,\mathcal{D}_q,\Psi_q)$.
Rather than instantiating one agent per subtask, \method groups
operationally compatible nodes into responsibility regions. We express this
organization by a surjective map:
\begin{equation}
\pi_q:\mathcal{U}_q\longrightarrow\mathcal{S}_q,
\qquad
\mathcal{U}_q=\coprod_{S\in\mathcal{S}_q}S,
\quad
\pi_q(u)=S\ \text{for }u\in S,
\label{eq:responsibility-map}
\end{equation}
where each $S\in\mathcal{S}_q$ is a responsibility region and the disjoint
union assigns every task node to exactly one such region. Contracting the
regions under $\pi_q$ yields the quotient topology:
\begin{equation}
\overline{H}_q
=H_q/\pi_q
=\left(\mathcal{S}_q,\overline{\mathcal{D}}_q\right),
\qquad
(S_i,S_j)\in\overline{\mathcal{D}}_q
\Longleftrightarrow
\exists (u,v)\in\mathcal{D}_q:
\pi_q(u)=S_i,\ \pi_q(v)=S_j,
\label{eq:responsibility-quotient}
\end{equation}
for $S_i\neq S_j$. Thus, $\overline{H}_q$ preserves dependencies between
regions while hiding dependencies internal to a single responsibility domain.
To preserve executability, a region merge is admissible only when the resulting
quotient remains acyclic.
To construct $\pi_q$, we associate each pair of task nodes with the
compatibility profile:
\begin{equation}
\mathbf{a}_q(u,v)
=\bigl(
a_{\Gamma}(u,v),
a_{\mathcal{Y}}(u,v),
a_{\mathcal{D}}(u,v),
a_{\mathrm{s}}(u,v)
\bigr),
\label{eq:task-affinity}
\end{equation}
whose components respectively measure tool overlap, input--output compatibility,
dependency adjacency, and semantic similarity. Starting from singleton regions,
operationally compatible regions are merged only when the resulting region
respects the prescribed task and tool capacities and the quotient remains
acyclic. Thus, $\pi_q$ uses complementary structural evidence without
introducing a separate free-form role decomposition.

\paragraph{Capability-grounded ownership.}
Let $\Gamma(S)=\bigcup_{u\in S}\Gamma(u)$ be the capabilities required by a
region. A responsibility assignment is a map:
\begin{equation}
\phi_q:\mathcal{S}_q\longrightarrow\mathcal{V}_q,
\qquad
\Gamma(S)\subseteq\Gamma_{\phi_q(S)}
\quad \text{for every }S\in\mathcal{S}_q,
\label{eq:capability-grounded-ownership}
\end{equation}
where $\mathcal{V}_q$ is the active agent set and $\Gamma_{A}$ denotes the
tools available to agent $A$. The map need not be injective: one agent may own
several compatible regions, while structurally or operationally distinct
regions remain assigned to different agents. If no existing agent covers a
region, the agent pool supplies a new specialization whose capabilities cover
$\Gamma(S)$. In this way, an agent role is defined by a responsibility domain
in the task topology rather than by a prefabricated textual label.

\subsection{Structure-Adaptive Collaborative Orchestration}
\label{sec:collaborative-orchestration}

The quotient topology determines which responsibility domains must exchange
information. Pushing cross-region task dependencies through the ownership map
gives the communication edges required for task execution:
\begin{equation}
\mathcal{E}^{\mathrm{req}}_q
=
\left\{
\bigl(\phi_q(S_i),\phi_q(S_j)\bigr)
\ \middle|\
 \exists (u,v)\in\mathcal{D}_q,\ u\in S_i,\ v\in S_j,
\ \phi_q(S_i)\neq\phi_q(S_j)
\right\}.
\label{eq:required-collaboration}
\end{equation}
The runtime collaboration topology combines these dependency-induced edges
with profile-conditioned spatial edges:
\begin{equation}
G_q=\left(
\mathcal{V}_q,
\mathcal{E}^{\mathrm{req}}_q\cup\mathcal{E}^{\mathrm{sp}}_q
\right).
\label{eq:collaboration-topology}
\end{equation}
Here $\mathcal{E}^{\mathrm{sp}}_q$ is constructed from agent-profile
similarity, query relevance, and tool input--output complementarity, providing
additional communication opportunities that may be pruned adaptively.

Let $z_i$ denote the artifact produced for region $S_i$. Local execution and
boundary-aware handoff are summarized by:
\begin{equation}
\begin{aligned}
z_i&=A_{\phi_q(S_i)}\!\left(
q|_{S_i},\Gamma(S_i),\mathcal{M}^{\mathrm{vis}}_q(S_i),\{h_{ji}\}
\right),\\
h_{ij}&=\operatorname{Route}_{ij}\!\left(z_i;\xi_{ij}\right)
\in\left\{\rho_{ij}(z_i),\varnothing\right\}.
\end{aligned}
\label{eq:regional-collaboration}
\end{equation}
Here $\rho_{ij}$ exposes only the artifact required by the downstream region,
whereas $\operatorname{Route}_{ij}$ applies the dynamic topology and routing
policy; a verifier request may be triggered when required.

\subsection{Boundary-Preserving Memory Coordination}
\label{sec:memory-coordination}

\method associates each memory record with its producing agent, task region,
and provenance. The responsibility topology induces a visibility policy
$\Lambda_q$ that determines which private records and admitted boundary evidence
are available to a region:
\begin{equation}
\mathcal{M}^{\mathrm{vis}}_q(S_i)
=
\operatorname{Filter}_{\Lambda_q}
\left(\mathcal{M}_q;S_i\right).
\label{eq:visible-memory}
\end{equation}
The filter first enforces ownership and boundary permissions; semantic
retrieval subsequently ranks only the admissible records. Specifically,
semantic retrieval selects seeds from the admissible view, after which typed
graph expansion reconstructs the regional context:
\begin{equation}
\mathcal{R}_q(S_i)
={\operatorname{Expand}_{L}}\!\left(
\operatorname{Seed}_{k}\!\left(
q|_{S_i},\mathcal{M}^{\mathrm{vis}}_q(S_i)
\right);
\mathcal{M}^{\mathrm{vis}}_q(S_i)
\right),
\label{eq:boundary-restricted-retrieval}
\end{equation}
where $k$ is the number of retrieval seeds and {$L$ is the
expansion depth.}
Semantic relevance can therefore rank admissible memories but cannot override
their ownership or sharing boundary. Records may encode episodic, evidence,
skill, subgraph, and failure relations.

The association between experience and responsibility is preserved when agent
ownership changes. If region $S$ is reassigned, its private memory is
transported with the region:
\begin{equation}
\phi_q^{(t)}(S)\neq\phi_q^{(t+1)}(S)
\quad\Longrightarrow\quad
\mu_{S}^{t\rightarrow t+1}:
\mathcal{M}^{\circ,(t)}_q(S)\longrightarrow
\mathcal{M}^{\circ,(t+1)}_q(S),
\label{eq:memory-transport}
\end{equation}
where $\mu_{S}^{t\rightarrow t+1}$ retains provenance while updating access
authority. When task-region ownership changes, records supported by the
reassigned tasks are transferred with provenance and their access metadata is
updated. This prevents both orphaned experience and unintended global sharing.

\subsection{Coherence-Constrained Structural Evolution}
\label{sec:structural-evolution}

{The structural projection of the task-conditioned
configuration $C_q$ is:}
\begin{equation}
{\Omega_q=\operatorname{Struct}(C_q)
=\left(\pi_q,\phi_q,G_q,\Lambda_q\right).}
\label{eq:task-organization}
\end{equation}
Here $\Lambda_q$ is the memory admission, visibility, retrieval, and transport
policy induced by the responsibility quotient; concrete records
$\mathcal{M}_q$ are runtime state governed by $\Lambda_q$, not genes in the
evolutionary configuration. {Under the coherence requirement
above,}
$\Omega_q$ is coherent when its ownership, collaboration, and memory policy are supported by the quotient
topology $\overline{H}_q$. The construction in the preceding modules establishes
this relation through the shared maps $\pi_q$ and $\phi_q$, rather than through
independent structural penalties.

{At evolution step $t$, execution feedback $\tau_q^{(t)}$ and
accumulated experience $\mathcal{E}^{(t)}$ guide a structural proposal:}
\begin{equation}
\widetilde{\Omega}_q^{(t+1)}
=\operatorname{Propose}\!\left(
\Omega_q^{(t)},\tau_q^{(t)},\mathcal{E}^{(t)}
\right).
\label{eq:coherent-evolution}
\end{equation}
{Equation~\ref{eq:coherent-evolution} summarizes proposal at
the configuration level: evolution changes reusable structural policies,
which are compiled into a candidate organization for the current task.
A proposal may change only a subset of policies. If it changes task-region
ownership, compilation reconciles required handoffs and memory permissions
with the new boundaries; other proposals must also preserve those task
dependencies. Coupling therefore requires compatible relations, not a change
to every component on every step.}

{A single structural screen checks that each task node has one
capable owner and that required cross-region handoffs and memory visibility
respect the induced boundaries. A valid candidate is then executed;
\(\operatorname{Accept}\) denotes reward-based selection against its parent
using the prescribed margin:}
\begin{equation}
\Omega_q^{(t+1)}
=
\begin{cases}
\widetilde{\Omega}_q^{(t+1)}, &
\operatorname{Valid}(\widetilde{\Omega}_q^{(t+1)})
\land
\operatorname{Accept}(\widetilde{\Omega}_q^{(t+1)}),\\
\Omega_q^{(t)}, & \text{otherwise}.
\end{cases}
\label{eq:coherent-selection}
\end{equation}
{An accepted organization is available to subsequently launched
tasks, while execution traces inform later proposals.}

%% file: Sections/Experiments.tex
\section{Experiments}
\label{sec:experiments}

We evaluate whether topological coherence across task, agent, collaboration, and
memory structures improves task performance, verified workflow progress,
handoff reliability, and information isolation. We then disentangle the
contributions of the structural components and evolution, and examine parameter
sensitivity and token cost conditioned on achieved workflow progress.

\subsection{Experimental Setup}
\label{sec:experimental-setup}

\noindent\textbf{Benchmarks.}
We evaluate reasoning on \textbf{BBEH}~\citep{kazemi2025big}, stateful tool use
on \textbf{WorkBench}~\citep{styles2024workbench}, and repository repair on
\textbf{SWE-Bench-Verified}~\citep{jimenez2023swe}, following the general
benchmark selection of evolutionary MAS work~\citep{hu2026evomas}. We also use
\textbf{CoMemBench}~\citep{zhao2026comembench} to assess multi-agent workflows with executable dependency
graphs, typed handoffs, and paired clean and polluted instances. Its native
evaluators separate terminal success from verified progress, information
transfer, and isolation. Dataset details appear in Appendix~\ref{app:benchmarks}.

\noindent\textbf{Evaluation Metrics.}
We report BBEH accuracy, WorkBench task success, and the official
SWE-Bench-Verified resolved rate. For CoMemBench~\citep{zhao2026comembench},
SR measures terminal success, VNCR verified task-node progress, VHS accepted
mandatory handoffs, and ICS memory isolation under paired pollution,
conditional on clean-target success. Definitions and denominators appear in
Appendix~\ref{app:metrics}. Token cost includes execution, coordination,
memory, and recovery calls; infrastructure failures and incomplete telemetry
are reported separately.

\noindent\textbf{Backbone Models.}
The main comparison uses \textbf{Qwen3.8-27B} and
\textbf{DeepSeek-V4-Flash}; the separate large-model comparison uses
\textbf{GPT-6-luna}. Each run uses a homogeneous
task-agent backbone, with model, tool schema, context budget, and decoding
policy controlled within each comparison.

\noindent\textbf{Baselines.}
We compare \textbf{Direct LLM Call} as a control; fixed organizations
\textbf{Peer Review}~\citep{xu2023towards} and
\textbf{SMoA}~\citep{li2025smoa}; automatic design methods
\textbf{ADAS}~\citep{hu2024automated} and
\textbf{OMAC}~\citep{li2026omac}; and the self-evolving baseline
\textbf{EvoMAS}~\citep{hu2026evomas}. Benchmark adapters record execution and
handoff evidence without changing a method's routing or memory decisions.

\noindent\textbf{Implementation Details.}
Linux clients access the backbones through OpenAI-compatible APIs; Qwen3.8-27B
inference runs on A100 GPUs, while CoMemBench execution uses H800 and A100
workers. Methods share benchmark tasks, tools, and native evaluators; paired
CoMemBench variants start from the same source state in isolated workspaces.
The fixed \method configuration uses organization affinity $0.45$ and memory
retrieval $k=3$; we vary these over $\{0.30,0.45,0.60,0.75\}$ and
$\{1,3,5,8\}$, respectively. In online structural evolution, a task's first
attempt uses the genome current at launch and supplies its reported score;
feedback then proposes one structural candidate, tested on that same task.
Accepted changes affect subsequently launched tasks, not the scored attempt.
We retain native evaluator receipts, handoff traces, and model-call ledgers for
auditing and cost accounting.

\subsection{Main Results}
\label{sec:main-results}

\begin{figure*}[t]
  \centering
  \includegraphics[width=0.96\textwidth]{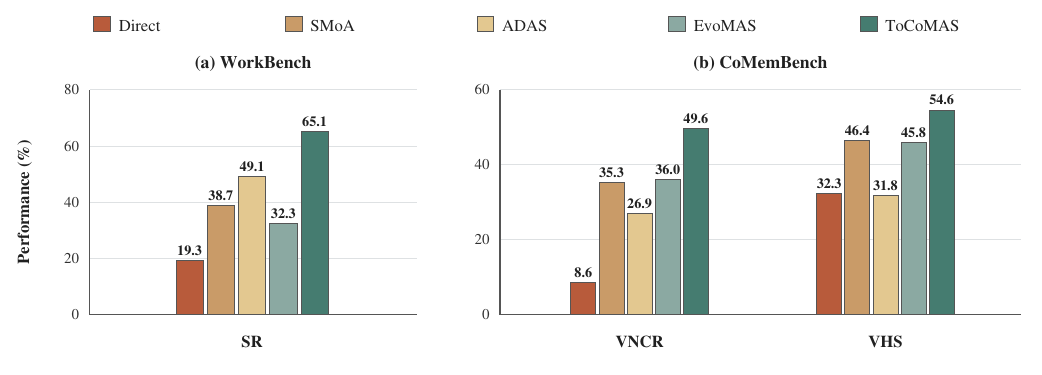}
  \caption{GPT-6-luna results on WorkBench SR and CoMemBench VNCR/VHS.}
  \label{fig:large-llm-results}
\end{figure*}

Table~\ref{tab:main-results} jointly reports general task performance and the
four CoMemBench quality measures for the Qwen3.8-27B and DeepSeek-V4-Flash
backbones. Figure~\ref{fig:large-llm-results}
reports the GPT-6-luna comparison. We draw two observations from the reported
results:

\input{tables/main_results}

\noindent\textbf{Topology-coherent organization improves general task
performance.} With Qwen3.8-27B, \method reaches 54.78\% on BBEH, 77.68\% on
WorkBench, and 48.00\% on SWE-Bench-Verified, exceeding the strongest reported
baseline by 8.26, 15.51, and 8.00 percentage points, respectively. With
DeepSeek-V4-Flash, the corresponding scores are {55.20}\%, 61.45\%, and 49.00\%,
also leading the reported baselines. These gains are consistent with grounding
agent responsibility and tool access in task regions and routing execution
across their dependency boundaries.

\noindent\textbf{The gains extend to verified handoffs and information
isolation.} On Qwen3.8-27B CoMemBench, \method attains 14.30\% SR, 41.60\%
VNCR, 51.00\% VHS, and {84.90}\% ICS, compared with EvoMAS's 1.00\%, 30.42\%,
30.73\%, and 26.80\%. Improvement is visible not only at the terminal outcome
but also in verified node progress, accepted cross-agent artifacts, and
resistance to polluted context. This pattern is consistent with deriving
collaboration interfaces and memory visibility from the same responsibility
topology; component contributions are examined in the ablations below. With
GPT-6-luna, \method reaches 65.07\% WorkBench SR and 49.58\% CoMemBench VNCR,
{and also leads EvoMAS on VHS (54.57\% versus 45.80\%).} CoMemBench SR is
9.00\% for \method and at most 0.50\% for the compared baselines.

\subsection{Ablation Study}
\label{sec:ablation}

\input{tables/ablation}

\begin{figure*}[t]
  \centering
  \begin{subfigure}[t]{0.45\textwidth}
    \centering
    \includegraphics[width=\linewidth]{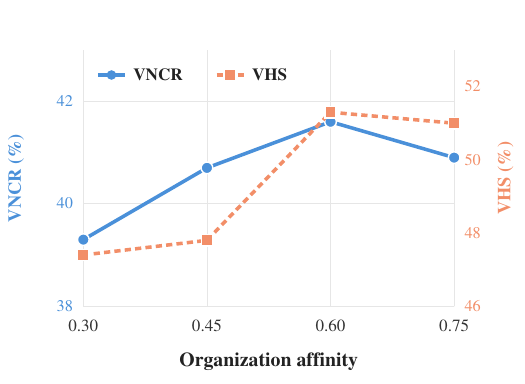}
    \caption{Organization affinity.}
  \end{subfigure}\hspace{0.04\textwidth}
  \begin{subfigure}[t]{0.45\textwidth}
    \centering
    \includegraphics[width=\linewidth]{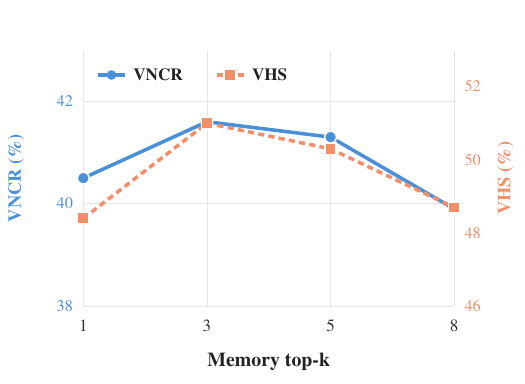}
    \caption{Memory retrieval budget.}
  \end{subfigure}
  \caption{Parameter sensitivity of ToCoMAS.}
  \label{fig:parameter-sensitivity}
  \vspace{-1em}
\end{figure*}

\noindent Table~\ref{tab:ablation} compares full \method with variants that
remove one structural mechanism at a time. \textbf{Topological Agent
Organization} assigns connected, tool-compatible task regions to agents;
replacing this assignment with canonical-order groups preserves tool coverage
but lowers CoMemBench VNCR by 14.33 points. This suggests that the placement of
responsibility matters beyond the availability of tools. \textbf{Structure-Adaptive
Collaborative Orchestration} derives inter-agent handoffs from cross-region task
dependencies and checks whether downstream agents can consume the projected
artifacts. Replacing these interfaces with fixed predecessor handoffs reduces
VHS by 40.85 points and CoMemBench SR to zero, exposing the cost of a broken
consumer boundary even when upstream outputs are verified. \textbf{Boundary-Preserving
Memory Coordination} restricts retrieval to private records and evidence
admitted across responsibility boundaries; making every verified record
globally visible reduces {ICS by 19.96 points}. Thus, record
validity alone does not determine whether it belongs in an agent's context.
Finally, \textbf{Coherence-Constrained Structural Evolution} uses task feedback
to update the organization for subsequent tasks; retaining a fixed structure
reduces WorkBench SR by 6.52 points. {Together, these ablations
link the three coherence relations to distinct outcomes: responsibility
assignment to verified progress, cross-region handoffs to artifact acceptance,
and memory boundaries to isolation; online evolution further improves task
success.}

\subsection{Parameter Sensitivity}
\label{sec:parameter-sensitivity}

Figure~\ref{fig:parameter-sensitivity} varies two controls while holding the
remaining settings fixed. The organization-affinity threshold governs how
readily related task nodes form an agent region. Both VNCR and VHS improve from
$0.30$ to $0.60$, where they reach $41.6\%$ and $51.3\%$, before declining at
$0.75$. Too little selectivity can mix weakly related responsibilities, while
too much can fragment useful regions. The memory retrieval budget $k$ limits
the records supplied to a task node. Both measures peak at $k=3$ and decline
as retrieval expands to $k=8$, consistent with extra context offering no
guaranteed benefit once relevant evidence is available. Together, the sweeps
favor a bounded task-region and memory scope rather than maximal grouping or
retrieval.

\subsection{Cost Analysis}
\label{sec:cost-analysis}

\begin{figure*}[t]
  \centering
  \includegraphics[width=0.9\textwidth,trim=0 0 0 22pt,clip]{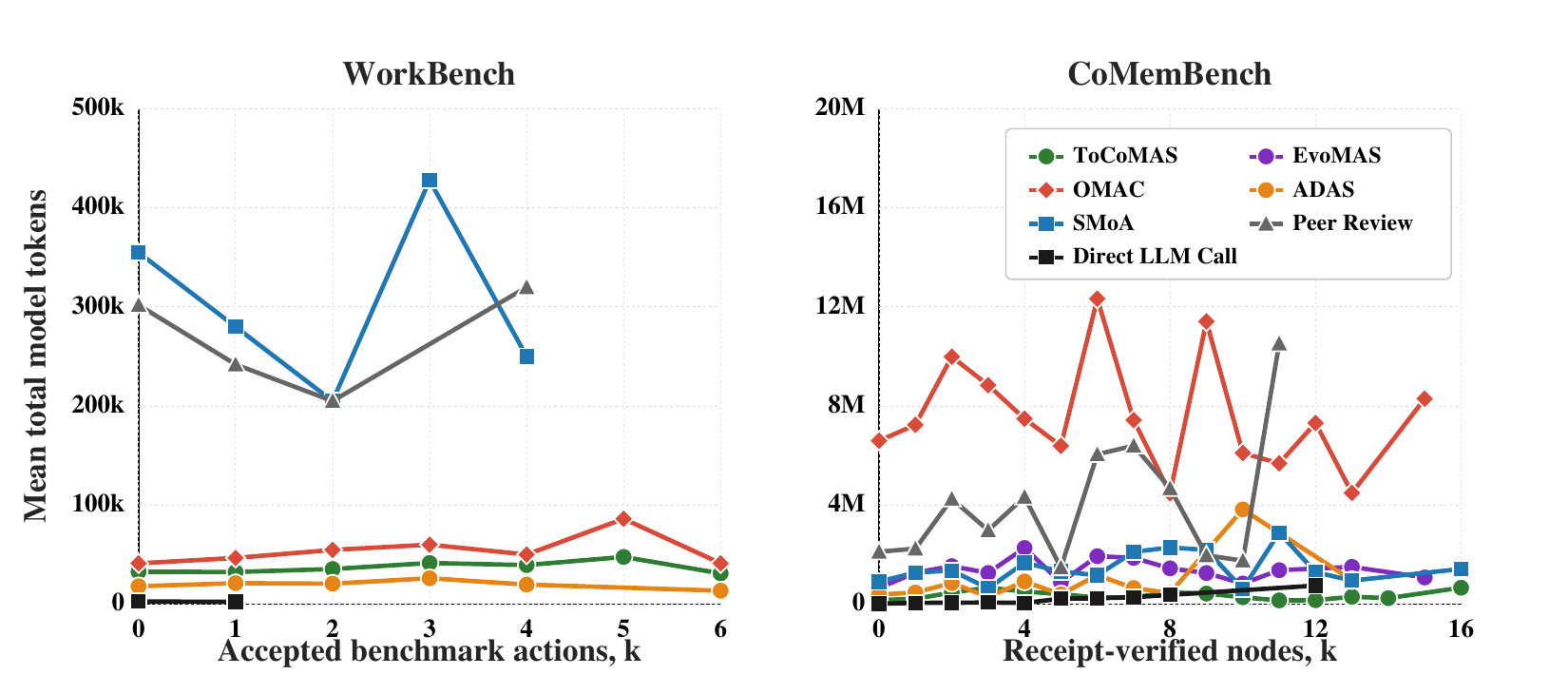}
  \caption{Progress-conditioned token cost across WorkBench and CoMemBench.}
  \label{fig:cost-analysis}
  \vspace{-1em}
\end{figure*}

We examine token use within bins of observed progress rather than
averaging over tasks of unequal length. For task $q$, let $n(q)$ be the number
of evaluator-matched actions on WorkBench or receipt-verified nodes on
CoMemBench, and let $c(q)$ be its recorded input and output model tokens.
Figure~\ref{fig:cost-analysis} plots the mean $c(q)$ within each observed
$n(q)=k$ bin, using successful WorkBench tasks and valid CoMemBench cells with
complete token telemetry. The WorkBench \method curve pairs post-recheck
progress with first-pass tokens; the CoMemBench panel uses a separate cohort.

On WorkBench, the plotted \method curve lies below Peer Review and SMoA but
above Direct LLM Call and ADAS. On CoMemBench, it lies below the other
multi-agent curves at most shared progress levels, although Direct LLM Call
remains cheaper. This pattern is consistent with the organization in
Figure~\ref{fig:motivation}: grouping compatible task nodes internalizes their
dependencies within an agent, so required handoffs arise only at cross-region
edges; projected artifacts and boundary-filtered retrieval limit the context
passed to downstream agents. Aligning these decisions with the task topology
can therefore reduce coordination overhead without suppressing verified
progress.

%% file: tables/main_results.tex
\begin{table*}[t]
  \centering
  \caption{Overall comparison on BBEH, WorkBench, SWE-Bench-Verified, and
  CoMemBench.}
  \label{tab:main-results}
  \scriptsize
  \setlength{\tabcolsep}{2.4pt}
  \renewcommand{\arraystretch}{1.05}
  \resizebox{\textwidth}{!}{%
  \begin{tabular}{@{}cllccc|cccc@{}}
    \toprule
    \multirow{2}{*}{\textbf{Backbone}} &
    \multirow{2}{*}{\textbf{Organization}} &
    \multirow{2}{*}{\textbf{Method}} &
    \multicolumn{3}{c|}{\textbf{General Task Performance}} &
    \multicolumn{4}{c}{\textbf{CoMemBench}} \\
    \cmidrule(lr){4-6}\cmidrule(lr){7-10}
    & & & \textbf{BBEH} $\uparrow$ & \textbf{WorkBench} $\uparrow$ &
    \textbf{SWE-V} $\uparrow$ & \textbf{SR} $\uparrow$ &
    \textbf{VNCR} $\uparrow$ & \textbf{VHS} $\uparrow$ &
    \textbf{ICS} $\uparrow$ \\
    \midrule

    \multirow{7}{*}{\textbf{Qwen3.8-27B}}
      & \multirow{1}{*}{Control} & Direct LLM Call & 19.13 & 19.28 & 9.80 & 0.00 & 8.81 & 34.50 & 0.00 \\
    \cmidrule(lr){2-10}
      & \multirow{2}{*}{Prespecified} & Peer Review & 46.52 & 37.10 & 30.50 & 1.00 & 30.74 & 33.37 & 19.10 \\
      & & SMoA & 37.17 & 37.25 & 35.00 & 4.00 & 38.73 & 43.32 & 10.10 \\
    \cmidrule(lr){2-10}
      & \multirow{2}{*}{Automatically designed} & ADAS & 32.61 & 62.17 & 34.20 & 0.00 & 13.91 & 27.73 & 7.40 \\
      & & OMAC & 41.09 & 55.94 & 39.50 & 2.00 & 32.98 & 40.51 & 17.30 \\
    \cmidrule(lr){2-10}
      & Self-evolving & EvoMAS & 37.60 & 59.13 & 40.00 & 1.00 & 30.42 & 30.73 & 26.80 \\
      & Ours & \cellcolor{blue!10}\method & \cellcolor{blue!10}54.78 & \cellcolor{blue!10}77.68 & \cellcolor{blue!10}48.00 & \cellcolor{blue!10}14.30 & \cellcolor{blue!10}41.60 & \cellcolor{blue!10}51.00 & \cellcolor{blue!10}84.90 \\
    \midrule

    \multirow{7}{*}{\textbf{DeepSeek-V4-Flash}}
      & \multirow{1}{*}{Control} & Direct LLM Call & 14.13 & 20.43 & 12.00 & 0.0 & 9.03 & 27.54 & 6.50 \\
    \cmidrule(lr){2-10}
      & \multirow{2}{*}{Prespecified} & Peer Review & 49.35 & 39.57 & 39.60 & 0.50 & 27.51 & 29.45 & 17.10 \\
      & & SMoA & 43.91 & 44.06 & 43.00 & 1.00 & 37.92 & 41.96 & 14.90 \\
    \cmidrule(lr){2-10}
      & \multirow{2}{*}{Automatically designed} & ADAS & 30.65 & 52.03 & 34.20 & 3.00 & 33.84 & 36.17 & 14.30 \\
      & & OMAC & 41.30 & 35.36 & 43.00 & 3.50 & 56.88 & 55.61 & 4.40 \\
    \cmidrule(lr){2-10}
      & Self-evolving & EvoMAS & 50.65 &55.50 & 46.00 & 1.50 & 27.99 & 27.17 & 9.80 \\
      & Ours & \cellcolor{blue!10}\method & \cellcolor{blue!10}55.20 & \cellcolor{blue!10}61.45 & \cellcolor{blue!10}49.00 & \cellcolor{blue!10}14.00 & \cellcolor{blue!10}44.36 & \cellcolor{blue!10}59.92 & \cellcolor{blue!10}76.50 \\
    \bottomrule
  \end{tabular}%
  }
  \vspace{-1em}
\end{table*}

%% file: tables/ablation.tex
\begin{table*}[t]
  \centering
  \caption{Architectural ablations and fixed-versus-evolving comparison of \method.}
  \label{tab:ablation}
  \scriptsize
  \setlength{\tabcolsep}{5.0pt}
  \begin{tabular}{lccccc}
    \toprule
    \multirow{2}{*}{\textbf{Variant}} & \multicolumn{1}{c}{\textbf{WorkBench}} &
    \multicolumn{4}{c}{\textbf{CoMemBench}} \\
    \cmidrule(lr){2-2}\cmidrule(lr){3-6}
    & \textbf{SR} $\uparrow$ & \textbf{SR} $\uparrow$ &
    \textbf{VNCR} $\uparrow$ & \textbf{VHS} $\uparrow$ & \textbf{{ICS}} $\uparrow$ \\
    \midrule
    w/o Topological Agent Organization & 69.86 & 4.00 & 27.27 & 37.97 & 67.14 \\
    w/o Structure-Adaptive Collaborative Orchestration & 67.25 & 0.00 & 28.20 & 10.15 & 66.70 \\
    w/o Boundary-Preserving Memory Coordination & 68.55 & 6.50 & 36.30 & 42.07 & 64.94 \\
    w/o Coherence-Constrained Structural Evolution & 71.16 & 10.60 & 40.07 & 47.80 & 80.80 \\
    \rowcolor{blue!10}\method & 77.68 & 14.30 & 41.60 & 51.00 & 84.90 \\
    \bottomrule
  \end{tabular}
  \vspace{-1em}
\end{table*}

%% file: Sections/Conclusion.tex
\section{Conclusion}

In this work, we presented \method, a topology-coherent architecture for
self-evolving multi-agent systems. Rather than specifying agent roles,
communication, and memory independently, \method grounds task nodes in tool
requirements, groups compatible regions into agent responsibilities, and
derives handoffs and memory boundaries from the resulting organization.
{Online evolution compiles structural proposals into
task-conditioned organizations and retains improvements that satisfy
executable and coherence constraints.} Experiments on BBEH, WorkBench, and SWE-Bench-Verified
show improved task performance over the evaluated baselines; CoMemBench further
reveals {gains over the self-evolving baseline} in verified node completion, accepted handoffs, and isolation
from polluted context. Ablations identify distinct contributions from agent
organization, collaborative orchestration, memory coordination, and structural
evolution. These results support topological coherence as a guiding principle
for organizing and evolving multi-agent systems.

%% file: Sections/Reproducibility.tex
Experimental settings are detailed in Section~\ref{sec:experimental-setup}.
Native evaluators and retained execution receipts, handoff traces, and
model-call ledgers support auditing.

%% file: Sections/AIUse.tex
Generative AI tools assisted with manuscript drafting and language editing.
The authors reviewed AI-assisted material and remain responsible for the
scientific claims and reported results.

%% file: Sections/Appendix.tex
\section{Benchmark and Metric Details}
\label{app:benchmarks}

\subsection{Benchmark Instances}

\textbf{BBEH}~\citep{kazemi2025big} tests difficult reasoning questions against
benchmark answers. \textbf{WorkBench}~\citep{styles2024workbench} tests workplace
tasks involving application tools. \textbf{SWE-Bench-Verified}~\citep{jimenez2023swe}
tests repository patches against its official resolution criterion. We use
each benchmark's native evaluator.

\textbf{CoMemBench}~\citep{zhao2026comembench} provides executable workflows in
four domains: stateful tool use, repository code, offline retrieval, and formal
mathematics. It contains 800 clean workflows (200 per domain) and 800 matched
polluted variants. A workflow specifies a request,
canonical task nodes and dependency edges, node-local execution requirements,
typed handoff contracts, resources, and a domain-native evaluator. Dependency
edges represent required task relations, not predefined agent communication
edges. Verified node receipts and artifact-exposure records support the
progress and handoff metrics below. In a matched polluted variant, one
contextually inapplicable artifact is added to a target consumer's incoming
artifact view; the query, graph, initial state, tools, and non-target inputs
remain unchanged.

\subsection{Metric Definitions and Aggregation}
\label{app:metrics}

For BBEH, WorkBench, and SWE-Bench-Verified, we report their respective
benchmark-level metrics: answer accuracy, task success rate, and official
resolved rate. The definitions below apply specifically to CoMemBench.

For CoMemBench, let $\mathcal{C}$ be the evaluated clean workflows. The
canonical required nodes of workflow $q$ are $\mathcal{U}_q$, consistent with
$T_q$ in Section~\ref{sec:method}; let
$\mathcal{U}_q^{\mathrm{ver}}\subseteq\mathcal{U}_q$ contain nodes with valid native-verifier
receipts. Terminal \emph{Success Rate} (SR) and \emph{Verified Node Completion
Rate} (VNCR) are:
\begin{equation}
  \mathrm{SR}
  =\frac{1}{|\mathcal{C}|}
   \sum_{q\in\mathcal{C}}\mathbb{I}[\operatorname{terminal}(q)=1],
  \qquad
  \mathrm{VNCR}
  =\frac{1}{|\mathcal{C}|}
   \sum_{q\in\mathcal{C}}
   \frac{|\mathcal{U}_q^{\mathrm{ver}}|}{|\mathcal{U}_q|}.
  \label{eq:app-sr-vncr}
\end{equation}
VNCR averages within-workflow fractions, so workflows receive equal weight;
method-private nodes and repeated attempts at a canonical node do not add to
the numerator.

Let $\mathcal{H}$ contain each non-root consumer whose mandatory direct
predecessors have verified. A consumer belongs to $\mathcal{H}^{+}$ only if
all required predecessor artifacts are exposed at the declared boundary,
current, receipt-valid, explicitly consumed, and followed by native-verifier
success. \emph{Verified Handoff Success} (VHS) measures required sharing:
\begin{equation}
  \mathrm{VHS}=\frac{|\mathcal{H}^{+}|}{|\mathcal{H}|}.
  \label{eq:app-vhs}
\end{equation}

For isolation, each matched pair $z$ has a target consumer $t_z$. Let
$\mathcal{P}$ contain pairs whose target verifies in the clean run, and let
$\mathcal{P}^{+}$ contain those pairs whose target also verifies after the
inapplicable artifact is introduced in the polluted run. \emph{Isolation
Challenge Success} (ICS) measures the preservation of a valid target outcome
under this memory-boundary intervention:
\begin{equation}
  \mathrm{ICS}=\frac{|\mathcal{P}^{+}|}{|\mathcal{P}|},
  \qquad
  \mathcal{P}=\bigcup_{d\in\mathcal{D}}\mathcal{P}_d,
  \label{eq:app-ics}
\end{equation}
where $\mathcal{D}$ is the set of four CoMemBench domains. The aggregate is
micro-pooled over eligible target pairs, not averaged over domain percentages;
eligibility requires clean \emph{target} success, not full clean-workflow
success. An eligible polluted target that is not reached or fails verification
does not enter $\mathcal{P}^{+}$. ICS does not require an explicit rejection
message. For VHS or ICS, a cohort with no eligible cases is reported as
unavailable rather than zero; infrastructure failures are tracked separately
from method outcomes. All four CoMemBench ratios are multiplied by 100 when
reported as percentages.